\documentclass[conference]{IEEEtran}
\IEEEoverridecommandlockouts
\usepackage{balance}
\usepackage{hyperref} 
\usepackage{stfloats}
\usepackage{amsmath}
\usepackage{caption}
\usepackage{booktabs}
\usepackage{amsmath,amssymb,amsfonts}
\usepackage{algorithmic}
\usepackage{graphicx}
\usepackage{textcomp}
\usepackage{xcolor}
\def\BibTeX{{\rm B\kern-.05em{\sc i\kern-.025em b}\kern-.08em
    T\kern-.1667em\lower.7ex\hbox{E}\kern-.125emX}}
\begin{document}

\title{Localization-Aware Multi-Scale Representation Learning for Repetitive Action Counting
\thanks{$^{\ast}$ Equal contribution. $^{\dagger}$ indicates the corresponding author. This work was supported in part by the National Key Research and Development Program of China under Grant 2023YFF1105101 and 62306031.}
}
\author{Sujia Wang$^{1,\ast}$, Xiangwei Shen$^{1,\ast}$, Yansong Tang$^{1}$, Xin Dong$^{1}$, Wenjia Geng$^{1}$, Lei Chen$^{2,\dagger}$   \\$^1$Shenzhen International Graduate School, Tsinghua University $^2$Department of Automation, Tsinghua University \\
    \{wangsujia0103, xwshen2022\}@gmail.com, \{tang.yansong\}@sz.tsinghua.edu.cn, \{leichenthu\}@tsinghua.edu.cn
}
\makeatletter
\let\@oldmaketitle\@maketitle
\renewcommand{\@maketitle}{\@oldmaketitle
\setcounter{figure}{0} 
\begin{minipage}{\textwidth}
\centering
\includegraphics[width=\linewidth]{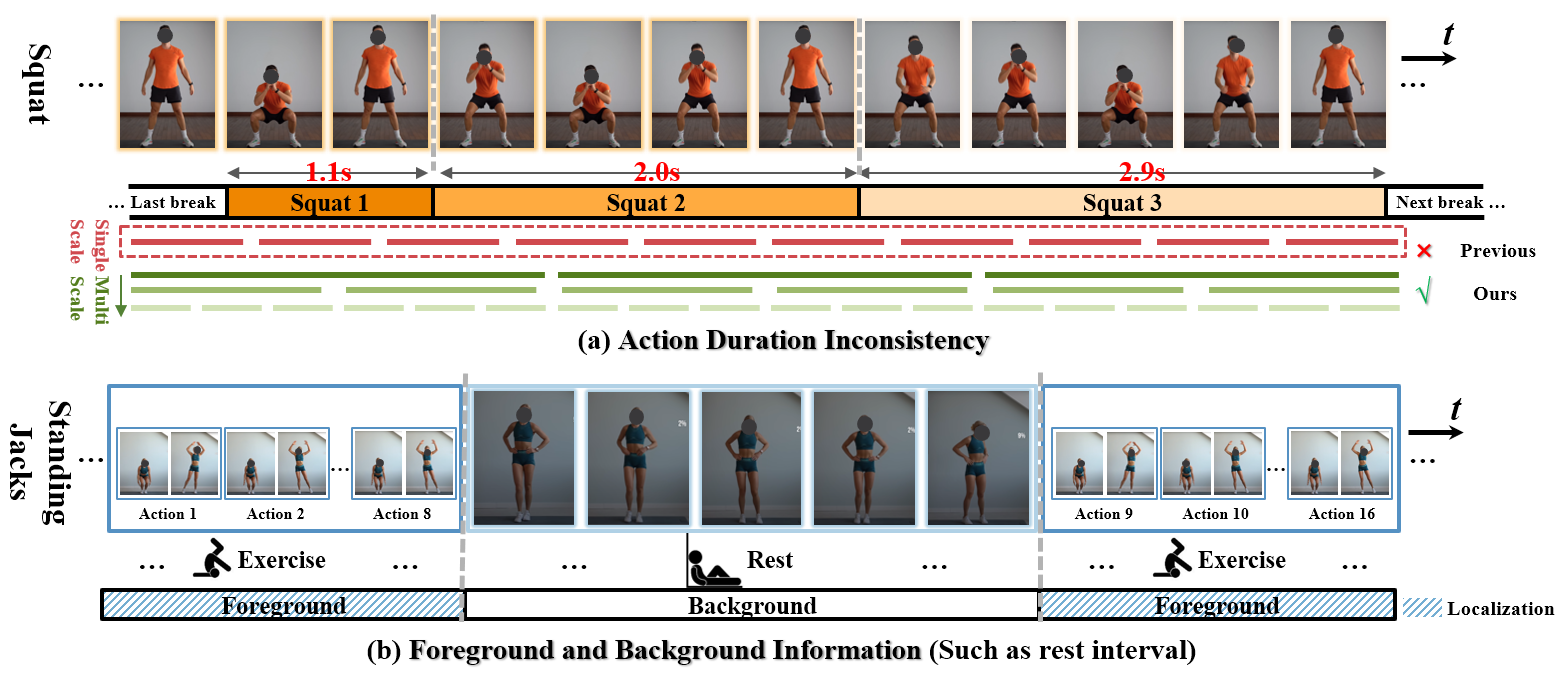}
    \captionof{figure}{\textbf{Two challenging cases existing in Repetitive Action Counting (face masked)}. (a) A man squats slower as physical strength declines. The action duration often changes, leading to failure prediction in previous methods due to their single-scale representation. In contrast, we propose a new Localization-Aware-Multi-Scale Representation Learning approach to handle frequency variation. (b) A woman rests between exercise sets. Our proposed methods refine the representation by localizing the foreground interests, shown as continuous repetitive exercising action blocks.}
    \label{fig:teaser}
\end{minipage}}
\makeatother
\maketitle
\begin{abstract}
Repetitive action counting (RAC) aims to estimate the number of class-agnostic action occurrences in a video without exemplars. Most current RAC methods rely on a raw frame-to-frame similarity representation for period prediction. However, this approach can be significantly disrupted by common noise such as action interruptions and inconsistencies, leading to sub-optimal counting performance in realistic scenarios. In this paper, we introduce a foreground localization optimization objective into similarity representation learning to obtain more robust and efficient video features. We propose a Localization-Aware Multi-Scale Representation Learning (LMRL) framework. Specifically, we apply a Multi-Scale Period-Aware Representation (MPR) with a scale-specific design to accommodate various action frequencies and learn more flexible temporal correlations. Furthermore, we introduce the Repetition Foreground Localization (RFL) method, which enhances the representation by coarsely identifying periodic actions and incorporating global semantic information. These two modules can be jointly optimized, resulting in a more discerning periodic action representation. Our approach significantly reduces the impact of noise, thereby improving counting accuracy. Additionally, the framework is designed to be scalable and adaptable to different types of video content. Experimental results on the RepCountA and UCFRep datasets demonstrate that our proposed method effectively handles repetitive action counting.
\end{abstract}
\begin{IEEEkeywords}
Repetitive action counting, Multi-scale representation, Foreground localization, Learnable similarity matrix.

\end{IEEEkeywords}
\vspace{-5pt}
\section{Introduction}
Repetitive actions are a fundamental part of daily life, such as stirring soup or chopping vegetables. Identifying and counting these actions in videos is essential for systems that observe and understand human activities over time. Repetitive action counting (\textbf{RAC}) aims to count class-agnostic repetitive actions in videos. Despite substantial efforts, existing temporal self-similarity-based methods underperform in diverse action scenes. Fig. \ref{fig:teaser} illustrates two real-life scenarios of action inconsistency and interruptions, which current methods often overlook. The first example shows a man squatting at varying speeds as his strength declines, and the second example depicts a woman taking a short break during leg raises. Such inconsistencies and interruptions can degrade the performance of periodic action prediction.

Early methods \cite{levy, pogalin} focused on counting in limited scenarios, such as synthetic noise squares rotating periodically, but struggled with high variability. Recent datasets and methods \cite{zhang2020context, repnet, hu2022transrac} have advanced RAC, but action duration inconsistencies still affect temporal self-similarity representations, leading to sub-optimal period prediction. Hu et al. \cite{hu2022transrac} addressed this with multi-scale strategies and improved temporal resolution, but faced challenges with action interruptions, as shown in Fig. \ref{fig:teaser}.

We propose a Localization-Aware Multi-Scale Representation Learning approach for periodic action representations. Our method uses scale-specific attention to distinguish periodic actions from background noise. Additionally, we incorporate coarse-grained dense action localization to identify periodic actions at the instance level using global semantic information and action interval annotations. These modules are jointly optimized, yielding a fine-grained periodic action representation.

\section{Related Works}
\label{sec:related}
\textit{\textbf{(1) Repetitive Counting. }}The early work of \textbf{RAC} focused on class-agnostic counting (CAC) task in images. Lu et al. \cite{countingclass} first counts specific objects that are rarely seen in the training set. Subsequent works \cite{countingclass, countingfully, countingobject, countingwild, shi2022represent} have explored CAC, adopting few-shot inference as a standard experimental setup.
Recently, \textbf{RAC} has gained attention due to the abundance of human repetitive action videos. Early methods by Levy et al. \cite{levy} and Pogalin et al. \cite{pogalin} relied on periodic assumptions but were limited by small, uniform frequency datasets \cite{runia2018real, levy}, restricting their applicability to real-world scenarios. Dwibedi et al. \cite{repnet} proposed the RepNet framework using a temporal self-similarity matrix for periodic prediction. While effective for repetitive counting, these methods perform poorly in noisy, realistic environments due to their inability to handle variability. A robust representation that addresses foreground recognition and varying action durations is urgently needed.

\textit{\textbf{(2) Temporal Self-Similarity Matrix. }}In RAC, The Temporal Self-similarity Matrix (TSM) is inspired by spatial attention mechanisms, such as cross-attention for multi-modality information injection \cite{p2p, latent-diff} and spatial self-attention for pixel-level information fusion \cite{nonlocal, vaswani2017attention}. TSM is widely used in action recognition \cite{junejo2010view,junejo2008cross, kwon2021learning, cao2021few} due to its effective period modeling. In RAC, Dwibedi et al. \cite{repnet} first utilized TSM as a periodic intermediate representation for class-agnostic periodic prediction. They calculated pairwise L2 distances of pooled temporal embeddings to capture cycle information from temporal series. Despite its simplicity and generalizability, this method is sensitive to action duration inconsistencies. Recent works have attempted to address this by introducing fine-grained temporal priors. Zhang et al. \cite{zhang2020context} proposed an auto-regressive method to estimate cycle length, and Hu et al. \cite{hu2022transrac} introduced multi-scale temporal sampling on video embeddings. Unlike these approaches, we emphasize that the multi-scale representation of TSM itself is crucial for robust repetition perception.

\vspace{-8pt}
\section{Method}
Given a realistic video with random breaks and varying action durations, our goal is to quantify the frequency of recurring actions. To achieve this, we propose the Localization-Aware Multi-Scale Representation Learning (LMRL) framework, which comprises three components: Video Feature Extractor, Periodic Representation, and Period Predictor. This section first outlines the overall paradigm of our RAC framework, followed by a detailed introduction to the proposed repetition representation module.

\vspace{-10pt}
\subsection{Repetitive Action Counting Framework} RAC aims to output the number of class-agnostic repetitive actions $Y \in \mathbb{N}$ of a given video $F=\{f_i\}_{1}^{N} \in \mathbb{R}^{(C\times H \times W)\times N}$, where N stands for the video length and $(C\times H \times W)$ means the spatial channel. The overview of our repetitive action counting framework is described as follows: 

\textit{(1) Video Feature Extractor.} The first part of the paradigm is a pre-trained video backbone to extract the action representation from the original video \cite{videoswin}. We first partition video frames into consecutive subsequences using a 4-frame sliding window with a step size of 1, which can be defined as:
        \begin{equation}
            V = \{\{f_1,...f_4\};\{f_2,...,f_5\};...\}
        \end{equation}
        Then, these sequences $V$ are fed as the input to the Video Swin Transformer \cite{videoswin} and output embeddings $X = \{x_i\}_1^N \in \mathbb{R}^{N\times C}, x_i \in \mathbb{R}^{C}$. $x_i$ represents the embeddings of a single RGB frame and $C$ represents the number of the channels.
            
\textit{(2) Periodic Representation.} With embeddings $X$ extracted from the video backbone, we proposed a novel representation learning method, which consists of \textbf{MPR} and \textbf{RFL} modules to catch the action's repetitive critical information to reinforce the video's feature. Through the module, we can get the output reinforced embeddings $X' = \{x'_i\}_1^N \in \mathbb{R}^{N\times  C^{'}}, x_i \in \mathbb{R}^{ C^{'}}$.   
        
\textit{(3) Period Predictor.} Following previous works \cite{liu2019context,wan2019adaptive,ranjan2018iterative,hu2022transrac}, we adopt a density map predictor that receipts embeddings $X'$ representing a periodic pattern and outputs corresponding density map. Here, we use a transformer \cite{vaswani2017attention} to predict the density map $D$.



\begin{figure}
    \centering
    \includegraphics[width=1\linewidth]{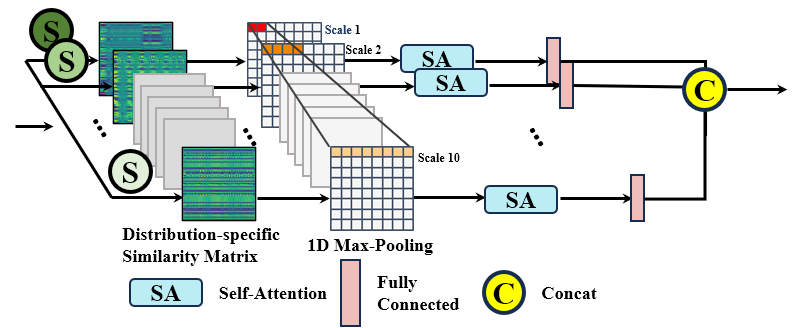}
    \caption{\textbf{Multi-Scale Period-Aware Representation. }}
    \label{fig:multi_pipe}
    \vspace{-10pt}
\end{figure}
\subsection{Multi-Scale Period-Aware Representation (MPR)} In this network, we devise a multi-scale architecture to catch patterns from different scale-specific aspects. 
The Multi-Scale Period-Aware Similarity Representation \textbf{(MPR)} is constructed hierarchically with different scales. For each scale, there are two main components: the Distribution-Specific Similarity Matrix and Scale-Specific Attention.

\textit{(1) Distribution-Specific Similarity Matrix.}
This module assigns varying weights to the temporal frame-to-frame similarity matrix to align with diverse action distributions. Specifically, we compute a similarity matrix \( S \) by calculating the pairwise similarity between embeddings \( x_i \) and \( x_j \) as \( S_{ij} = x_i^{T}Wx_j \in \mathbb{R}^{N \times N} \), where \( W \) is a learnable weight. We then apply a \( 1 \times 1 \) convolution layer followed by a dilated convolution with ReLU activation to enhance temporal information interaction.

\textit{(2) Scale-Specific Attention.} We employ a multi-scale strategy aimed at capturing diverse patterns of different lengths as shown in Fig. \ref{fig:multi_pipe}. For each scale, the similarity map passes through a Max-Pooling layer with distinct kernel size allowing the extraction of information with specific receptive fields. Given a $N \times N$ similarity map $S$, the pooling layer selects the max signal from $S$ with a sliding window with a step size of $2\times k_i$ and a kernel size of $2\times k_i$, where $N$ denotes video length and $k_i$ is the order number of scale. After that, we apply a self-attention mechanism following \cite{vaswani2017attention}  to better interact with temporal information. 
Then, we use a fully connected layer to get a scale vector with a size of $N\times 1$ by taking the weighted average of the attention channel. After that, we concatenate vectors from each scale to get $P \in \mathbb{R}^{N\times K}$ which models a new repetitive pattern, where $K$ is the number of scales. 

\begin{figure}
    \centering
    \includegraphics[width=0.9\linewidth]{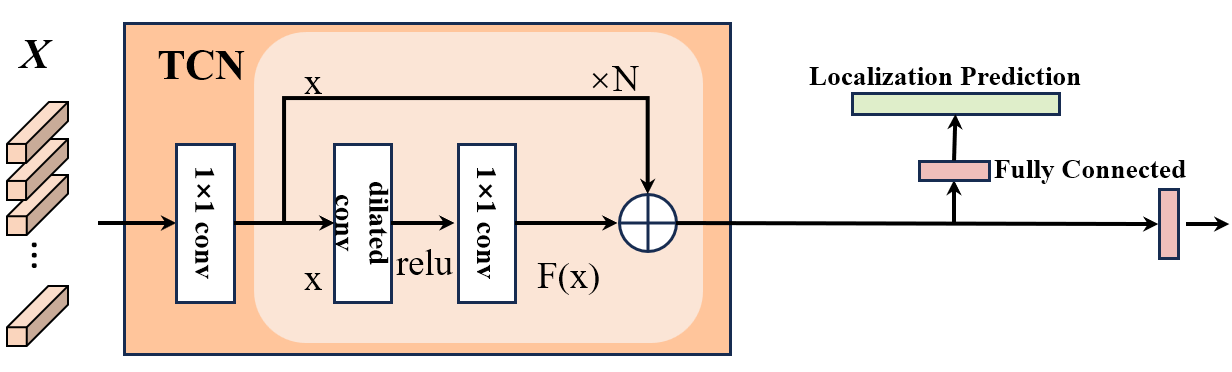}
    \caption{\textbf{Detailed RFL branch.}}
    \label{fig:localization}
    \vspace{-15pt}
\end{figure}
\vspace{-5pt}
\subsection{Repetition Foreground Localization (RFL)} The objective of the Repetition Foreground Localization (\textbf{RFL}) module is to localize the action cycle amidst noisy information and capture contextual information from feature sequences at the instance level. To facilitate direct supervision for noise detection, we add a fully connected layer to its outputs for intermediate prediction and localization optimization. The TCNs consist of a \(1 \times 1\) convolution to adjust input dimensions and \(n\) dilated residual blocks. Each block includes a dilated convolution layer with a large receptive field to capture long-range context information, followed by ReLU activation and a \(1 \times 1\) convolution layer. 

\textit{(1) Integration.} Given a feature sequence \( V \) from the backbone, we input it into both the MPR and RFL branches. Each branch applies a fully connected layer to adjust the output channel dimensions to \(  C^{'}/2 \). We then concatenate these output embeddings along the channel dimension and pass them through a LayerNorm and ReLU layer, resulting in a final \( N \times C^{'} \) embeddings for the period predictor.

\label{subsec:loss}

\textit{(2) Losses.} Following \cite{hu2022transrac}, we employ the Mean Absolute Error loss and Mean Squared Error loss to supervise the density map and the predicted count from the period predictor respectively. The overall counting loss is defined as follows:
\begin{equation}
     \mathcal{L}_{cou}= \frac{\left | c-\hat{c} \right | }{\hat{c} }+\alpha\cdot \frac{1}{N}\sum_{j=1}^{N} (y_j - \hat{y}_j)^2
\end{equation}
where $\hat{c}$ denotes the ground truth count, $\hat{y}$ represents the ground truth density at the $j$-th frame, and $B$ represents the batch size. Existing methods only supervise the period predictor, treating features from all periods equally in their correlation matrices. We argue that an ideal correlation matrix should exhibit high similarity scores during the acting period and low scores for unrelated periods to minimize noise influence. To achieve this, we introduce two auxiliary supervisions. These supervisory signals require no extra labeling and can be derived from existing dataset annotations.  Specifically, frames outside the action cycle's start and end are labeled as background (0), while the rest are foreground (1). Following \cite{li2020ms}, we apply binary cross-entropy loss with a smooth loss to the intermediate prediction:
\begin{equation}
    \mathcal{L}_{\text{loc}} = \frac{1}{T} \sum_{t=1}^{N} -\log \left( y_{t, c} \right) +  \frac{1}{2N} \sum_{t=1}^{N} \sum_{c} \left(y_{t-1, c} - y_{t, c}\right)^2
\end{equation}

where \( y_{t,0} \) and \( y_{t,1} \) denote the predicted probabilities of the background and foreground periods at instants \( t \), respectively, \( \tilde{y_{t}} \) denotes the ground truth class. In the MPR branch, we employ triplet margin loss widely used in metric learning and contrastive learning methods, which aims at discriminating the background period and reinforce the feature representation by clustering the embeddings of repetitive action periods while pushing away embeddings of the background period:
\begin{equation}
    \mathcal{L} _{tri} = \max\left\{ d\left(a_{i},p_{i} \right ) -d\left (a_{i},n_{i} \right ) + margin,0 \right \} 
\end{equation}
where $ d\left ( a,b  \right ) $ is the L2 distance between features, and $a,p,n$ are anchor, foreground and background samples.

\section{Experiment}
\begin{figure}
\vspace{-10pt}
\centering
    \includegraphics[width=\linewidth]{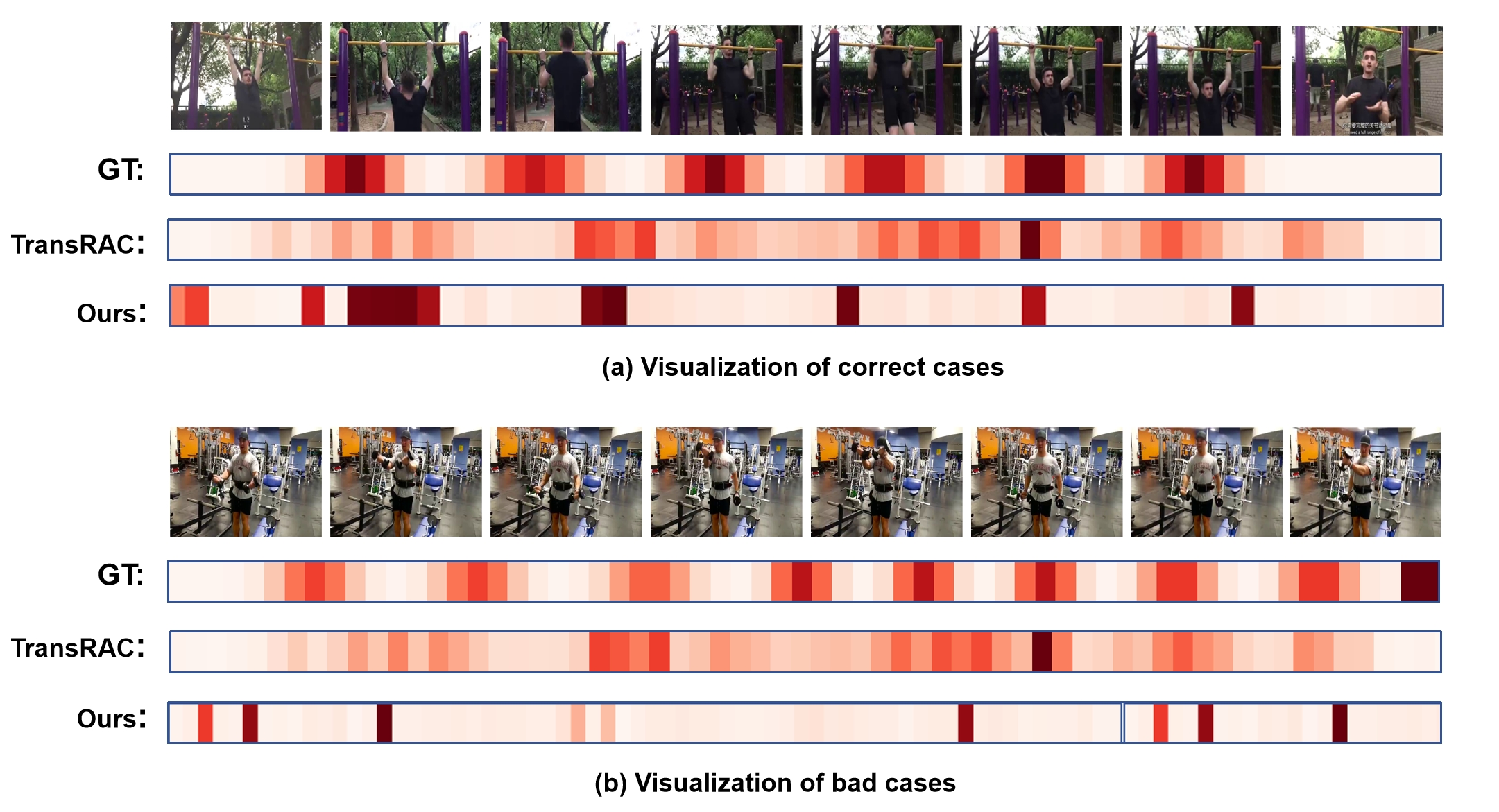}
    \caption{\textbf{Visualization of density map prediction.} The intensity of red color indicates the level of action density.}
    \label{fig:visual}
    \vspace{-10pt} 
\end{figure}

\begin{table}
\centering
\caption{\textbf{Performance of different methods on RepCount and UCFRep when trained on RepCount.}}
\renewcommand\tabcolsep{3.1mm}
\begin{tabular}{l|ll|cc}
\hline
             & \multicolumn{2}{l|}{RepCountA}        & \multicolumn{2}{l}{UCFRep}                            \\ \hline
             & \multicolumn{1}{l|}{$ \mathrm{MAE} \downarrow$}    & $\mathrm{OBO}\uparrow$    & \multicolumn{1}{l|}{$ \mathrm{MAE} \downarrow$}    & \multicolumn{1}{l}{$\mathrm{OBO}\uparrow$} \\ \hline
X3D\cite{feichtenhofer2020x3d}          & \multicolumn{1}{l|}{0.9105} & 0.1059 & \multicolumn{1}{c|}{-}      & -                       \\
TANet\cite{liu2021tam}        & \multicolumn{1}{l|}{0.6624} & 0.0993 & \multicolumn{1}{c|}{-}      & -                       \\
SwinT\cite{videoswin}        & \multicolumn{1}{l|}{0.5756} & 0.1324 & \multicolumn{1}{c|}{-}      & -                       \\
Huang et al\cite{huang2020improving}. & \multicolumn{1}{l|}{0.5267} & 0.1589 & \multicolumn{1}{c|}{-}      & -                       \\ \hline
RepNet\cite{repnet}       & \multicolumn{1}{l|}{0.9950} & 0.0134 & \multicolumn{1}{c|}{0.9985} & 0.009                   \\
Zhang et al\cite{zhang2020context} & \multicolumn{1}{l|}{0.8786}          & 0.1554          & \multicolumn{1}{c|}{0.7492} & 0.3802          \\
TransRAC\cite{hu2022transrac}     & \multicolumn{1}{l|}{0.4431} & 0.2913 & \multicolumn{1}{c|}{0.6402} & 0.324                   \\
LMRL (Ours)        & \multicolumn{1}{l|}{\textbf{0.4200}} & \textbf{0.3816} & \multicolumn{1}{c|}{0.7124} & \textbf{0.3954} \\ \hline
\end{tabular}
\label{table:main}
\end{table}
\begin{table}[t]
\centering
\renewcommand\tabcolsep{3.8mm}
\caption{\textbf{Experiments on integrating in our framework}. $ c$ and  $f$ refer to RFL and MPR, $\sigma \left ( c,f  \right ) $ for weighted linear average and $+$ for channel-wise concatenation.}
\begin{tabular}{l|cc|cc}
\hline
            & \multicolumn{2}{c|}{RepCountA}       & \multicolumn{2}{c}{UCFRep}           \\ \hline
Method & \multicolumn{1}{c|}{$ \mathrm{MAE} \downarrow$}    & $\mathrm{OBO}\uparrow$   & \multicolumn{1}{c|}{$ \mathrm{MAE} \downarrow$}    & $\mathrm{OBO}\uparrow$   \\ \hline
 $c$& \multicolumn{1}{c|}{0.4783} & 0.2895 & \multicolumn{1}{c|}{0.8786} & 0.3346 \\
$f $& \multicolumn{1}{c|}{0.5089} & 0.3092 & \multicolumn{1}{c|}{0.8202} & 0.3650 \\
$\sigma \left ( c,f  \right ) $ & \multicolumn{1}{c|}{\textbf{0.4200}} & \textbf{0.3816} & \multicolumn{1}{c|}{\textbf{0.7125}} & \textbf{0.3954} \\
$c+f$& \multicolumn{1}{c|}{0.4982} & 0.3552 & \multicolumn{1}{c|}{0.7458} & 0.3935 \\ \hline
\end{tabular}
%

\label{table:inter}

\end{table}

\begin{table}
\centering
\caption{\textbf{Ablations on applying different loss function}, where ${L}_{loc}$ indicates supervision on the RFL branch, $\mathcal{L}_{tri}$ for MPR branch and $\mathcal{L}_{den}$ refers to supervision on density map.}
\renewcommand\tabcolsep{1.7mm}
\begin{tabular}{@{}ccc|ccc|cc@{}}
\hline
\multicolumn{3}{c|}{\textbf{Loss}} & \multicolumn{3}{c|}{\textbf{Localization}}                              & \multicolumn{2}{c}{\textbf{Counting}} \\ \midrule
$\mathcal{L}_{loc}$         & $\mathcal{L}_{tri}$         & $\mathcal{L}_{den}$       & \multicolumn{1}{c|}{Acc}  & \multicolumn{1}{c|}{Edit} & F1@{10,25,50}  & \multicolumn{1}{c|}{Mae}     & Obo    \\ \midrule
$\times$    & $\times$    & \checkmark & \multicolumn{1}{c|}{51.2} & \multicolumn{1}{c|}{29.6} & 22.3 16.1 8.8  & \multicolumn{1}{c|}{0.4565}  & 0.2566 \\
$\times$     &\checkmark & \checkmark & \multicolumn{1}{c|}{53.1} & \multicolumn{1}{c|}{30.9} & 22.7 16.1 8.6  & \multicolumn{1}{c|}{0.4418}  & 0.3092 \\
\checkmark  & $\times$    & \checkmark & \multicolumn{1}{c|}{76.9} & \multicolumn{1}{c|}{35.0} & 37.1 32.5 22.0 & \multicolumn{1}{c|}{0.4767}  & 0.2894 \\
\checkmark &
  \checkmark &
  $\times$ &
  \multicolumn{1}{c|}{77.8} &
  \multicolumn{1}{c|}{\textbf{35.3}} &
  38.1 34.0 23.7 &
  \multicolumn{1}{c|}{0.4330} &
  0.3223 \\
\checkmark &
  \checkmark &
  \checkmark &
  \multicolumn{1}{c|}{\textbf{77.9}} &
  \multicolumn{1}{c|}{34.4} &
  \textbf{37.9 34.8 25.8} &
  \multicolumn{1}{c|}{\textbf{0.4200}} &
  \textbf{0.3816} \\ 
  \hline
\end{tabular} 
\label{table:loss}
\end{table}

\begin{table}
\centering

\caption{\textbf{Ablations on applying different similarity matrix}, where SA and TSM refer to Self-Attention and temporal similarity matrix used in RepNet \cite{repnet}.}
\renewcommand\tabcolsep{2.4mm}
\renewcommand{\arraystretch}{1.1}
\begin{tabular}{c|llc|cc}
\hline
\textbf{} & \multicolumn{3}{c|}{\textbf{Localization}}                              & \multicolumn{2}{c}{\textbf{Counting}} \\ \hline
\textbf{Method} & \multicolumn{1}{c|}{Acc}  & \multicolumn{1}{c|}{Edit} & F1@{10,25,50}  & \multicolumn{1}{c|}{Mae}             & Obo             \\ \hline
TSM             & \multicolumn{1}{l|}{77.6} & \multicolumn{1}{l|}{31.6} & 37.3 34.5 26.8& \multicolumn{1}{c|}{0.5393}          & 0.2697\\
SA             & \multicolumn{1}{l|}{77.4} & \multicolumn{1}{l|}{33.2} & \textbf{38.4 35.0 27.6}& \multicolumn{1}{c|}{0.4737}          & 0.2960\\
MPR(TSM)       & \multicolumn{1}{l|}{77.9} & \multicolumn{1}{l|}{31.7} & 36.4 33.7 25.3& \multicolumn{1}{c|}{0.4471}  & 0.3289 \\
LMRL (Ours)                       & \multicolumn{1}{l|}{\textbf{77.9}} & \multicolumn{1}{l|}{\textbf{34.4}} & 38.0 34.9 25.8& \multicolumn{1}{c|}{\textbf{0.4200}} & \textbf{0.3816} \\ \hline
\end{tabular}

\label{table:similarity}
\end{table}

We evaluate our model on two recent and challenging datasets: RepCountA \cite{hu2022transrac}, UCFRep \cite{zhang2020context}. Both datasets provide fine-grained annotations for the beginning and end of each action period that can be transformed into annotations used in our supervision. Following \cite{zhang2020context,hu2022transrac,repnet}, we comprehensively evaluate our approach to RAC under two metrics, Off-By-One (OBO) count errors and Mean Absolute Error (MAE) to evaluate counting performance.
\label{subsec:compare}

\vspace{-5pt}
\subsection{Comparison with the State of the Art} We compare our method with various recent approaches for repetitive action counting and action recognition on the RepCountA and UCFRep datasets. We train the models on the training set of RepCountA and validate them on the test sets of RepCountA and UCFRep. Tab. \ref{table:main} shows that our LMRL approach outperforms state-of-the-art methods on the RepCountA dataset. Additionally, the results indicate that our model performs well across different datasets without the need for fine-tuning. As shown in Fig. \ref{fig:visual}, a failure case reveals that our model can sometimes misidentify foreground frames as inactive. We attribute this issue to the insufficiency of context information caused by the high sampling rate.
\vspace{-5pt}
\subsection{Ablation Study} In this section, we perform several ablations to verify the decisions made while designing our method. We train our models on the train set of RepCountA.

\textit{(1) Supervision.} We enhance our framework's localization capability by incorporating two auxiliary methods: binary cross-entropy loss and triplet margin loss. We compare the impact of each supervision method and evaluate model performance in localizing action periods using three widely employed metrics in action segmentation \cite{lea2017temporal, farha2019ms, liu2023diffusion, gao2021global2local}. We maintain the count loss supervision across all models to ensure accurate counting. Tab. \ref{table:loss} demonstrates the effect of each supervision method on the MPR module, showing that better localization models achieve superior counting results. Notably, the fourth row of Tab. \ref{table:loss} shows that our model achieves state-of-the-art performance even without density map prediction, highlighting the benefits of distinguishing foreground from background in the RAC task.

\textit{(2) Similarity Component.} In Tab. \ref{table:similarity}, we compare the performance of similarity setup in our model. Multi-head self-attention and the temporal self-similarity matrix (TSM) \cite{repnet} are common approaches for modeling frame correlations in earlier RAC methods. For TSM models, we conduct two ablations: first, substituting the distribution-specific similarity matrix with TSM, and second, replacing the entire MPR with TSM. For the self-attention model, we replace the MPR with self-attention, as our scale-specific attention requires a \( T \times T \) feature map. Our results indicate that both components of the MPR significantly improve counting performance and enhance localization capability. 


\textit{(3) Integration.} In Tab. \ref{table:inter}, we compare four approaches to integrate our framework. We observed that both branches fundamentally model repetitive actions. Our MPR exhibits strong cross-dataset performance, significantly enhanced by leveraging a joint representation from both components. This combination captures long-range temporal information with the RFL module and explores repetition via multi-scale similarity, demonstrating the efficiency of our bilateral design.

\vspace{-5pt}
\section{Conclusion}
This paper presents a novel Localization-Aware Multi-Scale Representation Learning (LMRL) approach to address repetitive action counting by modeling robust periodic patterns. To overcome action inconsistencies, we introduce a multi-scale period representation for scale-specific perception and a localization module to distinguish between foreground and background periods. These two branches are jointly optimized, resulting in a highly discerning action representation. In addition to repetitive action counting in general scenarios, our method has potential applications in domains such as kitchen activity recognition, where it can accurately track repetitive actions like chopping or stirring, enhancing the functionality of smart kitchen systems and activity monitoring.

\bibliographystyle{./IEEEtran}
\bibliography{main}
\end{document}